\documentclass{article}
\usepackage{multirow}
\usepackage{microtype}
\usepackage{graphicx}
\usepackage{subfigure}
\usepackage{booktabs} 

\usepackage{hyperref}

\usepackage[accepted]{icml2023}

\usepackage{amsmath}
\usepackage{amssymb}
\usepackage{mathtools}
\usepackage{amsthm}

\usepackage[capitalize,noabbrev]{cleveref}

\theoremstyle{plain}

\theoremstyle{definition}

\theoremstyle{remark}

\usepackage[textsize=tiny]{todonotes}

\icmltitlerunning{Cross-Architectural Positive Pairs improve the effectiveness of Self-Supervised Learning}

\begin{document}

\twocolumn[
\icmltitle{Cross-Architectural Positive Pairs improve the effectiveness of Self-Supervised Learning}



\icmlsetsymbol{equal}{*}

\begin{icmlauthorlist}
\icmlauthor{Pranav Singh}{yyy}
\icmlauthor{Jacopo Cirrone}{sch}
\end{icmlauthorlist}

\icmlaffiliation{yyy}{Department of Computer Science, Tandon School of Engineering, New York University,  New York, NY 11202, USA}
\icmlaffiliation{sch}{Center for Data Science, New York University, and Colton Center for Autoimmunity, NYU Grossman School of Medicine, New York, NY 10011, USA}

\icmlcorrespondingauthor{Pranav Singh}{ps4364@nyu.edu}

\icmlkeywords{Machine Learning, ICML}

\vskip 0.3in
]



\printAffiliationsAndNotice{}  

\begin{abstract}
Existing self-supervised techniques have extreme computational requirements and suffer a substantial drop in performance with a reduction in batch size or pretraining epochs. This paper presents Cross Architectural - Self Supervision (CASS), a novel self-supervised learning approach that leverages Transformer and CNN simultaneously. Compared to the existing state-of-the-art self-supervised learning approaches, we empirically show that CASS-trained CNNs and Transformers across four diverse datasets gained an average of 3.8\% with 1\% labeled data, 5.9\% with 10\% labeled data, and 10.13\% with 100\% labeled data while taking 69\% less time. We also show that CASS is much more robust to changes in batch size and training epochs than existing state-of-the-art self-supervised learning approaches. We have opensourced our code at \url{https://github.com/pranavsinghps1/CASS}.
\end{abstract}

\section{Introduction}
\label{intro}
Self-supervised learning has emerged as a powerful paradigm for learning representations that can be used for various downstream tasks like classification, object detection, and image segmentation. Pretraining with self-supervised techniques is label-free, allowing us to train even on unlabeled images. This is especially useful in fields with limited labeled data availability or if the cost and effort required to provide annotations are high. Medical Imaging is one field that can benefit from applying self-supervised techniques. 
Medical imaging is a field characterized by minimal data availability. First, data labeling typically requires domain-specific knowledge. Therefore, the requirement of large-scale clinical supervision may be cost and time prohibitive. Second, due to patient privacy, disease prevalence, and other limitations, it is often difficult to release imaging datasets for secondary analysis, research, and diagnosis. Third, due to an incomplete understanding of diseases. This could be either because the disease is emerging or because no mechanism is in place to systematically collect data about the prevalence and incidence of the disease. An example of the former is COVID-19 when despite collecting chest X-ray data spanning decades, the samples lacked data for COVID-19 \cite{sriram2021covid}. An example of the latter is autoimmune diseases. Statistically, autoimmune diseases affect 3\% of the US population or 9.9 million US citizens. There are still major outstanding research questions for autoimmune diseases regarding the presence of different cell types and their role in inflammation at the tissue level. The study of autoimmune diseases is critical because autoimmune diseases affect a large part of society and because these conditions have been on the rise recently \cite{galeotti2020autoimmune,lerner2015world,ehrenfeld2020covid}. Other fields like cancer and MRI image analysis have benefited from the application of artificial intelligence (AI). But for autoimmune diseases, the application of AI is particularly challenging due to minimal data availability, with the median dataset size for autoimmune diseases between 99-540 samples  \cite{tsakalidou2022computer, Stafford2020ASR}.

To overcome the limited availability of annotations, we turn to self-supervised learning. Models extract representations that can be fine-tuned even with a small amount of labeled data for various downstream tasks~\cite{sriram2021covid}. As a result, this learning approach avoids the relatively expensive and human-intensive task of data annotation. But self-supervised learning techniques suffer when limited data is available, especially in cases where the entire dataset size is smaller than the peak performing batch size for some of the leading self-supervised techniques. This calls for a reduction in the batch size; this again causes existing self-supervised techniques to drop performance; for example, state-of-the-art DINO \cite{caron2021emerging} drops classification performance by ~25\% when trained with batch size 8. Furthermore, existing self-supervised techniques are compute-intensive and trained using multiple GPU servers over multiple days. This makes them inaccessible to general practitioners.  


Existing approaches in the field of self-supervised learning rely purely on Convolutional Neural Networks (CNNs) or Transformers as the feature extraction backbone and learn feature representations by teaching the network to compare the extracted representations. Instead, we propose to combine a CNN and Transformer in a response-based contrastive method. In CASS, the extracted representations of each input image are compared across two branches representing each architecture (see Figure 1). By transferring features sensitive to translation equivariance and locality from CNN to Transformer, our proposed approach - CASS, learns more predictive data representations in limited data scenarios where a Transformer-only model cannot find them. We studied this quantitatively and qualitatively in Section \ref{results}. Our contributions are as follows:
\begin{itemize}
\item We introduce \textbf{C}ross \textbf{A}rchitectural  - \textbf{S}elf \textbf{S}upervision (CASS), a hybrid CNN-Transformer approach for learning improved data representations in a self-supervised setting in limited data availability problems in the medical image analysis domain \footnote{We have opensourced our code at \url{https://github.com/pranavsinghps1/CASS}}
\item We propose the use of CASS for analysis of autoimmune diseases such as dermatomyositis and demonstrate an improvement of 2.55\% 
compared to the existing state-of-the-art self-supervised approaches. To our knowledge, the autoimmune dataset contains 198 images and is the smallest known dataset for self-supervised learning.   
\item Since our focus is to study self-supervised techniques in the context of medical imaging. We evaluate CASS on three challenging medical image analysis problems (autoimmune disease cell classification, brain tumor classification, and skin lesion classification) on three public datasets (Dermofit Project Dataset \cite{Dermofit}, brain tumor MRI Dataset \cite{Cheng2017,s21062222} and ISIC 2019~\cite{Tschandl2018TheHD, Gutman2018SkinLA, Combalia2019BCN20000DL}) and find that CASS improves classification performance (F1 Score and Recall value) over the existing state of the art self-supervised techniques by an average of ~3.8\% using 1\% label fractions, 5.9 \% with 10\% label fractions and 10.13\% with 100\% label fractions.
\item Existing methods also suffer a severe drop in performance when trained for a reduced number of epochs or batch size (\cite{caron2021emerging, Grill2020BootstrapYO, chen2020simple}). We show that CASS is robust to these changes in Sections \ref{bs-var} and \ref{epoch-var}. 
\item New state-of-the-art self-supervised techniques often require significant computational requirements. This is a major hurdle as these methods can take around 20 GPU days to train \cite{Azizi2021BigSM}. This makes them inaccessible in limited computational resource settings. CASS, on average, takes 69\% less time than the existing state-of-the-art methods. We further expand on this result in Section 5.2.
\end{itemize}
\section{Background}

\subsection{Neural Network Architectures for Image Analysis}
\label{archs}
CNNs are a famous architecture of choice for many image analysis applications~\cite{khan2020survey}. CNNs learn more abstract visual concepts with a gradually increasing receptive field. They have two favorable inductive biases: (i) translation equivariance resulting in the ability to learn equally well with shifted object positions, and (ii) locality resulting in the ability to capture pixel-level closeness in the input data. CNNs have been used for many medical image analysis applications, such as disease diagnosis~\cite{yadav2019deep} or semantic segmentation~\cite{ronneberger2015u}. To address the requirement of additional context for a more holistic image understanding, the Vision Transformer (ViT) architecture \cite{dosovitskiy2020image} has been adapted to images from language-related tasks and recently gained popularity~\cite{liu2021Swin,liu2021swinv2,pmlr-v139-touvron21a}. In a ViT, the input image is split into patches that are treated as tokens in a self-attention mechanism. Compared to CNNs, ViTs can capture additional image context but lack ingrained inductive biases of translation and location. As a result, ViTs typically outperform CNNs on larger datasets~\cite{d2021convit}. 

\subsubsection{Cross-architecture Technqiues}
Cross-architecture techniques aim to combine the features of CNNs and Transformers; they can be classified into two categories (i) Hybrid cross architecture techniques and (ii) pure cross-architecture techniques. 
Hybrid cross-architecture techniques combine parts of CNNs and Transformers in some capacity, allowing architectures to learn unique representations. ConViT~\cite{d2021convit} combines CNNs and ViTs using gated positional self-attention (GPSA) to create a soft convolution similar to inductive bias and improve upon the
capabilities of Transformers alone. More recently, the training regimes and inferences from ViTs have been used to design a new family of convolutional architectures - ConvNext \cite{liu2022convnet}, outperforming benchmarks set by ViTs in classification tasks. \cite{li2021bossnas} further simplified the procedure to create an optimal CNN-Transformer using their self-supervised Neural Architecture Search (NAS) approach. 

On the other hand, pure cross-architecture techniques combine CNNs and Transformers without any changes to their architecture to help both of them learn better representations. \cite{Gong2022CMKDCC} used CNN and Transformer pairs in a consistent teaching knowledge distillation format for audio classification and showed that cross-architecture distillation makes distilled models less prone to overfitting and also improves robustness. Compared with the CNN-attention hybrid
models, cross-architecture knowledge distillation is more effective
and does not require any model architecture change. Similarly, \cite{guo2022cross} also used a 3D-CNN and Transformer to learn strong representations and proposed a self-supervised learning module to predict an edit distance between two video sequences in the temporal order. Although their approach showed encouraging results on two datasets, their approach relies on both positive and negative pairs. Furthermore, their proposed approach is batch statistic dependent. 

\subsection{Self-Supervised Learning}

Most existing self-supervised techniques can be classified into contrastive and reconstruction-based techniques. Traditionally, contrastive self-supervised techniques have been trained by reducing the distance between representations of different augmented views of the same image (‘positive pairs’) and increasing the distance between representations of augmented views from different images (‘negative pairs’) \cite {He2020MomentumCF,Chen2020ASF,caron2020unsupervised}. But this is highly memory intensive as we need to track positive and negative pairs. Recently, Bootstrap Your Own Latent (BYOL) \cite{Grill2020BootstrapYO} and DINO \cite{caron2021emerging} have improved upon this approach by eliminating the memory banks. The premise of using negative pairs is to avoid collapse. Several strategies have been developed with BYOL using a momentum encoder, Simple Siamese (SimSiam) \cite{Chen2021ExploringSS} a stop gradient, and DINO applying the counterbalancing effects of sharpening and centering on avoiding collapse. Techniques relying only on the positive pairs are much more efficient than the ones using positive and negative pairs. Recently, there has been a surge in reconstruction-based self-supervised pretraining methods with the introduction of MSN \cite{assran2022masked}, and MAE \cite{he2021masked}. These methods learn semantic knowledge of the image by masking a part of it and then predicting the masked portion. 

\subsubsection{Self-supervised Learning and Medical Image Analysis}
\label{ssl-med}
ImageNet is most commonly used for benchmarking and comparing self-supervised techniques. ImageNet is a balanced dataset that is not representative of real-world data, especially in the field of medical imaging, that has been characterized by class imbalance. Self-supervised methods that use batch-level statistics have been found to drop a significant amount of performance in image classification tasks when trained on ImageNet by artificially inducing class imbalance \cite{assran2022hidden}.
This prior of some self-supervised techniques like MSN \cite{assran2022masked}, SimCLR \cite{chen2020simple}, and VICreg \cite{bardes2021vicreg} limits their applicability on imbalanced datasets, especially in the case of medical imaging. 

Existing self-supervised techniques typically require large batch sizes and datasets. When these conditions are not met, a marked reduction in performance is demonstrated~\cite{caron2021emerging,chen2020simple, Caron2020UnsupervisedLO, Grill2020BootstrapYO}. Self-supervised learning approaches are practical in big data medical applications~\cite{ghesu2022self,azizi2021big}, such as analysis of dermatology and radiology imaging. In more limited data scenarios (3,662 images - 25,333 images), \citet{Matsoukas2021IsIT} reported that ViTs outperform their CNN counterparts when self-supervised pre-training is followed by supervised fine-tuning. Transfer learning favors ViTs when applying standard
training protocols and settings. Their study included running the DINO \cite{caron2021emerging} self-supervised method over 300 epochs with a batch size of 256. However, questions remain about the accuracy and
efficiency of using existing self-supervised techniques on datasets whose entire size is smaller than their peak performance batch size. Also, viewing this from the general practitioner's perspective with limited computational power raises the
question of how we can make practical self-supervised approaches more accessible. Adoption and faster development of self-supervised paradigms will only be possible when they become easy to plug and play with limited computational power.

In this work, we explore these questions by designing CASS, a novel self-supervised approach developed with the core values of efficiency and effectiveness. In simple terms, we are combining CNN and Transformer in a response-based contrastive method by reducing similarity to combine the abilities of CNNs and Transformers. This approach was initially designed for a 198-image dataset for muscle biopsies of inflammatory lesions from patients with dermatomyositis - an autoimmune disease. The benefits of this approach are illustrated by challenges in diagnosing autoimmune diseases due to their rarity, limited data availability, and heterogeneous features. Consequently, misdiagnoses are common, and the resulting diagnostic delay plays a significant factor in their high mortality rate. Autoimmune diseases share commonalities with COVID-19 regarding clinical manifestations, immune responses, and pathogenic mechanisms. Moreover, some patients have developed autoimmune diseases after COVID-19 infection \cite{Liu2020COVID19AA}. Despite this increasing prevalence, the representation of autoimmune diseases in medical imaging and deep learning is limited.

\section{Methodology}

We start by motivating our method before explaining it in detail (in Section \ref{desc-cass}). Self-supervised methods have been using different augmentations of the same image to create positive pairs. These were then passed through the same architectures but with a different set of parameters \cite{Grill2020BootstrapYO}. In \cite{caron2021emerging} the authors introduced image cropping of different sizes to add local and global information. They also used different operators and techniques to avoid collapse, as described in Section 2.2.\\

But there can be another way to create positive pairs - through architectural differences. 
\cite{Raghu2021DoVT} in their study suggested that for the same input, Transformers and CNNs extract different representations. They conducted their study by analyzing the CKA (Centered Kernel Alignment) for CNNs and Transformer using ResNet \cite{He2016DeepRL} and ViT (Vision Transformer) \cite{dosovitskiy2020image} family of encoders, respectively. They found that Transformers have a more uniform representation across all layers as compared to CNNs. They also have self-attention, enabling global information aggregation from shallow layers and skip connections that connect lower layers to higher layers, promising information transfer. Hence, lower and higher layers in Transformers show much more similarity than in CNNs. The receptive field of lower layers for Transformers is more extensive than in CNNs. While this receptive field gradually grows for CNNs, it becomes global for Transformers around the midway point. Transformers don't attend locally in their earlier layers, while CNNs do. Using local information earlier is essential for solid performance. CNNs have a more centered receptive field as opposed to a more globally spread receptive field of Transformers. Hence, representations drawn from the same input will differ for Transformers and CNNs. Until now, self-supervised techniques have used only one kind of architecture at a time, either a CNN or Transformer. But differences in the representations learned with CNN and Transformers inspired us to create positive pairs by different architectures or feature extractors rather than using a different set of augmentations. This, by design, avoids collapse as the two architectures will never give the exact representation as output. By contrasting their extracted features at the end, we hope to help the Transformer learn representations from CNN and vice versa. This should help both the architectures to learn better representations and learn from patterns that they would miss. We verify this by studying attention maps and feature maps from supervised and CASS-trained CNN and Transformers in Appendix \ref{attn-map-appendix} and Section \ref{abl-attn-maps}. We observed that CASS-trained CNN and Transformer were able to retain a lot more detail about the input image, which pure CNN and Transformers lacked.

\subsection{Description of CASS}
\label{desc-cass}
CASS' goal is to extract and learn representations in a self-supervised way. To achieve this, an image is passed through a common set of augmentations. The augmented image is then simultaneously passed through a CNN and Transformer to create positive pairs. The output logits from the CNN and Transformer are then used to find cosine similarity loss (equation \ref{loss_eq}). This is the same loss function as used in BYOL \cite{grill2020bootstrap}. Furthermore, the intuition of CASS is very similar to that of BYOL. In BYOL to avoid collapse to a trivial solution the target and the online arm are differently parameterized and an additional predictor is used with the online arm. They compared this setup to that of GANs where joint of optimization of both arms to a common value was impossible due to differences in the arms. Analogously, In CASS instead of using an additional MLP on top of one of the arms and differently parameterizing them, we use two fundamentally different architectures. Since the two architectures give different output representations as mentioned in \cite{Raghu2021DoVT}, the model doesn't collapse. Additionally, to avoid collapse we introduced a condition where if the outputs from the CNN and Transformer are the same, artificial noise sampled from a Gaussian distribution is added to the model outputs and thereby making the loss non-zero.  
We also report results for CASS using a different set of CNNs and Transformers in Appendix \ref{change-arch} and Section \ref{results}, and not a single case of the model collapse was registered.

\begin{equation}
\label{loss_eq}
\operatorname{loss} =2-2 \times
 \operatorname{F(R)} \times \operatorname{F(T)}
 \end{equation}
\begin{align*}
\text{where, }
\operatorname{F(x)}=\sum_{i=1}^{N} \left(\frac{x}{\left(\operatorname{max}\left(\|x\|_{2}\right), \epsilon\right)}\right)
\end{align*}

We use the same parameters for the optimizer and learning schedule for both architectures. We also use stochastic weigh averaging (SWA) \cite{Izmailov2018AveragingWL} with Adam optimizer and a learning rate of 1e-3. For the learning rate, we use a cosine schedule with a maximum of 16 iterations and a minimum value of 1e-6. ResNets are typically trained with Stochastic Gradient Descent (SGD) and our use of the Adam optimizer is quite unconventional. Furthermore, unlike existing self-supervised techniques there is no parameter sharing between the two architectures. 

We compare CASS against the state-of-the-art self-supervised technique DINO (DIstilation with NO labels). This choice was made based on two conditions (i) As already explained in Section \ref{ssl-med}, some self-supervised techniques use batch-level statistics that makes them less suitable for application on imbalanced datasets and imbalanced datasets are a feature of medical imaging. (ii) The self-supervised technique should be benchmarked for both CNNs and Transformers as both architectures have exciting properties and apriori, it is difficult to predict which architecture will perform better. 

\begin{figure}[!htb]
    \centering
    \centering  
    \includegraphics[width=0.8\linewidth]{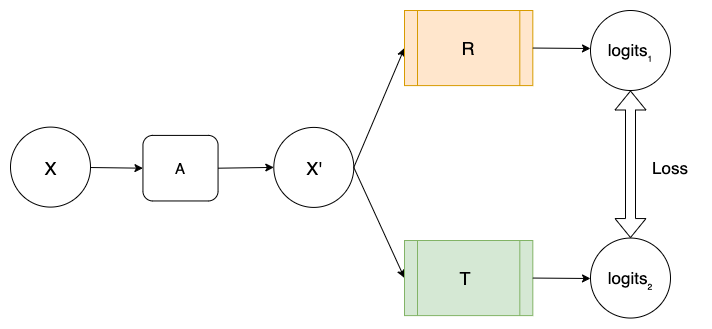}
    \hspace{1cm}
    \includegraphics[width=0.8\linewidth]{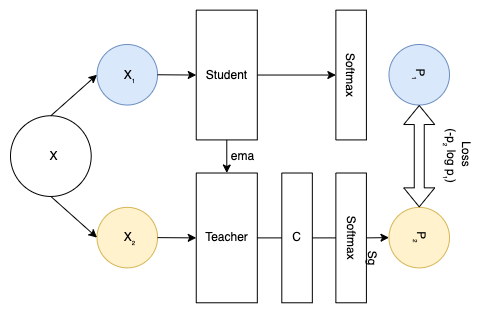}
    \caption{(Top) In our proposed self-supervised architecture - CASS, R represents ResNet-50, a CNN and T in the other box represents the Transformer used (ViT); X is the input image, which becomes X' after applying augmentations. Note that CASS applies only one set of augmentations to create X'. X' is passed through both arms to compute loss, as in Equation 1. This differs from DINO, which passes different augmentation of the same image through networks with the same architecture but different parameters. The output of the teacher network is centered on a mean computed over a batch. Another key difference is that in CASS, the loss is computed over logits; meanwhile, in DINO, it is computed over softmax output.}
\label{fig:CASS}
\end{figure}

In Figure~\ref{fig:CASS}, we show CASS on top, and DINO \cite{caron2021emerging} at the bottom. Comparing the two, CASS does not use any extra mathematical treatment used in DINO to avoid collapse such as centering and applying the softmax function on the output of its student and teacher networks. We also provide an ablation study using a softmax and sigmoid layer for CASS in Appendix B. After training CASS and DINO for one cycle, DINO yields only one kind of trained architecture. In contrast, CASS provides two trained architectures (1 - CNN and 1 - Transformer). CASS-pre-trained architectures perform better than DINO-pre-trained architectures in most cases, as further elaborated in Section \ref{results}. 

\section{Experimental Details}

\subsection{Datasets}

We split the datasets into three splits - training, validation, and testing following the 70/10/20 split strategy unless specified otherwise. We further expand upon our thought process for choosing datasets in Appendix \ref{ds-choice}.

\begin{table*}[!ht]
\centering
\begin{tabular}{llll}
\hline
\multicolumn{1}{c}{\multirow{2}{*}{Techniques}} & \multicolumn{1}{c}{\multirow{2}{*}{Backbone}} & \multicolumn{2}{l}{Testing F1 score} \\
\multicolumn{1}{c}{}                            & \multicolumn{1}{c}{}                           & 10\%       & 100\%         \\
\hline

DINO                                            & ResNet-50                                      & \textbf{0.8237±0.001}            &0.84252±0.008           \\
CASS                                           & ResNet-50                                   & 0.8158±0.0055           & \textbf{0.8650±0.0001}           \\
Supervised                                      & ResNet-50                                 & 0.819±0.0216          &  0.83895±0.007            \\
\hline

DINO                                            & ViT B/16                             & 0.8445±0.0008           &0.8639± 0.002             \\
CASS                                           & ViT B/16                                          & \textbf{0.8717±0.005}           & \textbf{0.8894±0.005}            \\
Supervised                                      & ViT B/16                                          &0.8356±0.007           &0.8420±0.009          \\
\hline
\end{tabular}
\caption{Results for autoimmune biopsy slides dataset. In this table, we compare the F1 score on the test set. We observed that CASS outperformed the existing state-of-art self-supervised method using 100\% labels for CNN as well as for Transformers. Although DINO outperforms CASS for CNN with 10\% labeled fraction. Overall, CASS outperforms DINO by ~2.2\% for 100\% labeled training for CNN and Transformer. For Transformers in 10\% labeled training CASS' performance was ~2.7\% better than DINO.}
\label{ai-perf}
\end{table*}

\begin{table}[!htb]
\begin{tabular}{lll}
\hline
Dataset    & DINO              & CASS                    \\
\hline
Autoimmune & 1 H 13 M    & \textbf{21 M}          \\
Dermofit & 3 H 9 M & \textbf{1 H 11 M} \\
Brain MRI  & 26 H 21 M  & \textbf{7 H 11 M}  \\
ISIC-2019  & 109 H 21 M & \textbf{29 H 58 M} \\
\hline
\end{tabular}
\caption{Self-supervised pretraining time comparison for 100 epochs on a single RTX8000 GPU. In this table, H represents hour(s), and M represents minute(s).}
\label{computetime}
\end{table}

\begin{itemize}

    \item \textbf{Autoimmune diseases biopsy slides} \cite{singh2022data,VANBUREN2022113233} consists of slides cut from muscle biopsies of dermatomyositis patients stained with different proteins and imaged to generate a dataset of 198 TIFF image set from 7 patients. The presence or absence of these cells helps to diagnose dermatomyositis. Multiple cell classes can be present per image; therefore this is a multi-label classification problem. 
    Our task here was to classify cells based on their protein staining into TFH-1, TFH-217, TFH-Like, B cells, and others. We used F1 score as our metric for evaluation, as employed in previous works by \cite{singh2022data, VANBUREN2022113233}. These RGB images have a consistent size of 352 by 469.
    
    \item \textbf{Dermofit dataset} \cite{Dermofit}  contains normal RGB images captured through an SLR camera indoors with ring lightning. There are 1300 image samples, classified into 10 classes: Actinic Keratosis (AK), Basal Cell Carcinoma (BCC), Melanocytic Nevus / Mole (ML), Squamous Cell Carcinoma (SCC), Seborrhoeic Keratosis (SK), Intraepithelial carcinoma (IEC), Pyogenic Granuloma (PYO), Haemangioma (VASC), Dermatofibroma (DF) and  Melanoma (MEL). This dataset comprises images of different sizes and no two images are of the same size. They range from 205×205 to 1020×1020 in size. Our pretext task is multi-class classification and we use the F1 score as our evaluation metric on this dataset. 
    
    \item \textbf{Brain tumor MRI dataset} \cite{Cheng2017, Amin2022ANM}  7022 images of human brain MRI that are classified into four classes: glioma, meningioma, no tumor, and pituitary. This dataset 
    combines Br35H: Brain tumor Detection 2020 dataset used in "Retrieval of Brain tumors by Adaptive Spatial Pooling and Fisher Vector Representation" and Brain tumor classification curated by Navoneel Chakrabarty and Swati Kanchan. Out of these, the dataset curator created the training and testing splits. We followed their splits, 5,712 images for training and 1,310 for testing. Since this was a combination of multiple datasets, the size of images varies throughout the dataset from 512×512 to 219×234. The pretext of the task is multi-class classification, and we used the F1 score as the metric. 
    
    \item \textbf{ISIC 2019} \cite{Tschandl2018TheHD, Gutman2018SkinLA, Combalia2019BCN20000DL} consists of 25,331 images across eight different categories - melanoma (MEL), melanocytic nevus (NV), Basal cell carcinoma (BCC), actinic keratosis(AK), benign keratosis(BKL), dermatofibroma(DF), vascular lesion (VASC) and Squamous cell carcinoma(SCC). This dataset contains images of size 600 × 450 and 1024 × 1024. The distribution of these labels is unbalanced across different classes. For evaluation, we followed the metric followed in the official competition i.e balanced multi-class accuracy value, which is semantically equal to recall. 
    \end{itemize}

\subsection{Self-supervised learning}
We studied and compared results between DINO and CASS-pre-trained self-supervised CNNs and Transformers. For the same, we trained from ImageNet initialization \cite{Matsoukas2021IsIT} for 100 epochs with a batch size of 16. We ran these experiments on an internal cluster with a single GPU unit (NVIDIA RTX8000) with 48 GB video RAM, 2 CPU cores, and 64 GB system RAM. 

For DINO, we used the hyperparameters and augmentations mentioned in the original implementation. For CASS, we describe the experimentation details in Appendix \ref{ssl-train}.
\subsection{End-to-end fine-tuning} In order to evaluate the utility of the learned representations, we use the self-supervised pre-trained weights for the downstream classification tasks. While performing the downstream fine-tuning, we perform the entire model (E2E
fine-tuning). The test set metrics were used as proxies for representation quality. We trained the entire model for a maximum of 50 epochs with an early stopping patience of 5 epochs. For supervised fine-tuning, we used Adam optimizer with a cosine annealing learning rate starting at 3e-04. Since almost all medical datasets have some class imbalance we applied class distribution normalized Focal Loss \cite{Lin2017FocalLF} to navigate class imbalance. 

We fine-tune the models using different label fractions during E2E fine-tuning i.e 1\%, 10\%, and 100\& label fractions. For example, if a model is trained with a 10\% label fraction, then that model will have access only to 10\% of the training dataset samples and their corresponding labels during the E2E
fine-tuning after initializing weights using the CASS or DINO pretraining. 


\section{Results and Discussion}
\label{results}

\subsection{Compute and Time analysis Analysis}
\label{time}
We ran all the experiments on a single NVIDIA
RTX8000 GPU with 48GB video memory. In Table \ref{computetime}, we compare the cumulative training times for self-supervised training of a CNN and Transformer with DINO and CASS. We observed that CASS took an average of 69\% less time compared to DINO. Another point to note is that CASS trained two architectures at the same time or in a single pass. While training a CNN and Transformer with DINO it would take two separate passes.

\subsection{Results on the four medical imaging datasets}

We did not perform 1\% finetuning for the autoimmune diseases biopsy slides of 198 images because using 1\% images would be too small a number to learn anything meaningful and the results would be highly randomized. Similarly, we also did not perform 1\% fine-tuning for the dermofit dataset as the training set was too small to draw meaningful results with just 10 samples. We present the results on the four medical imaging datasets in Tables \ref{ai-perf}, \ref{dermofitperformance}, \ref{brainMRIperformance}, and \ref{ISICperformance}. From these tables, we observe that CASS improves upon the classification performance of existing state-of-the-art self-supervised method DINO by 3.8\% with
1\% labeled data, 5.9\% with 10\% labeled data, and 10.13\% with 100\% labeled data.




\begin{table*}[!htb]
\centering
\begin{tabular}{lll}

\hline
\multicolumn{1}{c}{\multirow{2}{*}{Techniques}} & 
\multicolumn{2}{l}{Testing F1 score} \\
\multicolumn{1}{c}{}                                         & 10\%       & 100\%         \\
\hline

DINO                                             (Resnet-50)                                     &0.3749±0.0011          &0.6775±0.0005           \\
CASS                                            (Resnet-50)                                     & \textbf{0.4367±0.0002}      & \textbf{0.7132±0.0003}           \\
Supervised                                       (Resnet-50)                                   & 0.33±0.0001           &  0.6341±0.0077           \\
\hline

DINO                        (ViT B/16)                             &0.332± 0.0002             &0.4810±0.0012            \\
CASS    (ViT B/16)                                  & \textbf{0.3896±0.0013}           & \textbf{0.6667±0.0002}            \\
Supervised         (ViT B/16)                               &0.299±0.002          &0.456±0.0077          \\
\hline
\end{tabular}
\caption{This table contains the results for the dermofit dataset. We observe that CASS outperforms both supervised and existing state-of-the-art self-supervised methods for all label fractions. Parenthesis next to the techniques represents the architecture used, for example, DINO(ViT B/16) represents ViT B/16 trained with DINO. In this table, we compare the F1 score on the test set. We observed that CASS outperformed the existing state-of-art self-supervised method using all label fractions and for both the architectures.}
\label{dermofitperformance}
\end{table*}


\begin{table*}[t]
\centering
\begin{tabular}{lllll}
\hline
\multicolumn{1}{c}{\multirow{2}{*}{Techniques}} & \multicolumn{1}{c}{\multirow{2}{*}{Backbone}} & \multicolumn{3}{l}{Testing F1 score} \\
\multicolumn{1}{c}{}                            & \multicolumn{1}{c}{}                           & 1\%       & 10\%       & 100\%       \\
\hline
DINO                                            & Resnet-50                                      &\textbf{0.63405±0.09}          &\textbf{0.92325±0.02819}            & 0.9900±0.0058          \\
CASS                                           & Resnet-50                                      & 0.40816±0.13          & 0.8925±0.0254          &\textbf{0.9909±
0.0032}
             \\
Supervised                                      & Resnet-50                                      &0.52±0.018          &0.9022±0.011            & 0.9899± 0.003            \\
\hline
DINO                                            & ViT B/16                                          &0.3211±0.071      &0.7529±0.044           &0.8841±
0.0052
           \\
CASS                                           & ViT B/16                                           & \textbf{0.3345±0.11}          & \textbf{0.7833±0.0259}           &\textbf{0.9279± 
0.0213}
             \\
Supervised                                      & ViT B/16                                           & 0.3017 ± 0.077         & 0.747±0.0245           & 0.8719± 0.017           \\
\hline
\end{tabular}
\caption{This table contains results on the brain tumor MRI classification dataset. While DINO outperformed CASS for 1\% and 10\% labeled training for CNN, CASS maintained its superiority for 100\% labeled training, albeit by just 0.09\%. Similarly, CASS outperformed DINO for all data regimes for Transformers, incrementally 1.34\% in for 1\%, 3.04\% for 10\%, and 4.38\% for 100\% labeled training. We observe that this margin is more significant than for biopsy images. Such results could be ascribed to the increase in dataset size and increasing learnable information.}
\label{brainMRIperformance}
\end{table*}

\begin{table*}[!h]
\centering
\begin{tabular}{lllll}
\hline
\multicolumn{1}{c}{\multirow{2}{*}{Techniques}} & \multicolumn{1}{c}{\multirow{2}{*}{Backbone}} & \multicolumn{3}{l}{Testing Balanced multi-class accuracy} \\
\multicolumn{1}{c}{}                            & \multicolumn{1}{c}{}                           & 1\%       & 10\%       & 100\%       \\
\hline
DINO                                            & Resnet-50                                      &0.328±0.0016         &0.3797±0.0027            &0.493±3.9e-05            \\
CASS                                           & Resnet-50                                      &\textbf{0.3617±0.0047}            &\textbf{0.41±0.0019}            &  \textbf{0.543±2.85e-05}           \\
Supervised                                      & Resnet-50                                      &0.2640±0.031           &0.3070±0.0121            &0.35±0.006           \\ 
\hline
DINO                                            & ViT B/16                                           & 0.3676± 0.012           &0.3998±0.056          &0.5408±0.001            \\
CASS                                           & ViT B/16                                           &\textbf{0.3973± 0.0465}           &\textbf{0.4395±0.0179}            &  \textbf{0.5819±0.0015}          \\
Supervised                                      & ViT B/16                                           &0.3074±0.0005           & 0.3586±0.0314           &  0.42±0.007  \\
\hline
\end{tabular}

\caption{Results for the ISIC-2019  dataset. Comparable to the official metrics used in the challenge \url{https://challenge.isic-archive.com/landing/2019/}. The ISIC-2019 dataset is an incredibly challenging, not only because of the class imbalance issue but because it is made of partially processed and inconsistent images with hard-to-classify classes. We use balanced multi-class accuracy as our metric, which is semantically equal to recall value. We observed that CASS consistently outperforms DINO by approximately 4\% for all label fractions with CNN and Transformer.}
\label{ISICperformance}
\end{table*}

\begin{figure}[!h]
    \centering
    
    \includegraphics[width=0.7\linewidth]{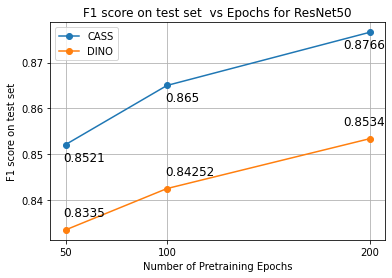}
    \hspace{1cm}
    \includegraphics[width=0.7\linewidth]{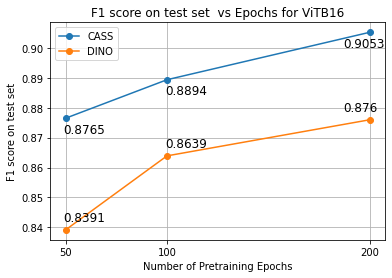}\\
    (a)
    

    \centering
    \centering  
    
    \includegraphics[width=0.7\linewidth]{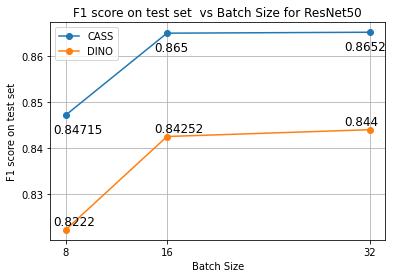}
    \hspace{1cm}
    \includegraphics[width=0.7\linewidth]{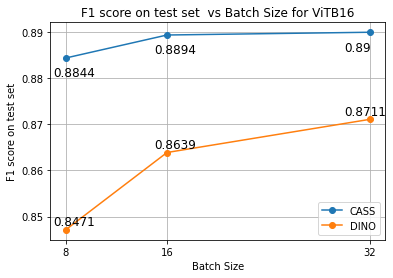}\\
    (b)
    
    \caption{In Figure a, we report the change in performance with respect to the change in the number of pretraining epochs for DINO and CASS for ResNet-50 and ViTB/16, respectively. In Figure b, we report the change in performance with respect to the change in the number of pretraining batch sizes for DINO and CASS for ResNet-50 and ViTB/16, respectively. These ablation studies were conducted on the autoimmune dataset, while keeping the other hyper-parameters the same during pretraining and downstream finetuning.}
\label{fig:batchsize}
\end{figure}

\subsection{Ablation Studies}

As mentioned in Section \ref{ssl-med}, existing self-supervised methods experience a drop in classification performance when trained for a reduced number of pretraining epochs and batch size. We performed ablation studies to study the effect of change in performance for CASS and DINO pre-trained ResNet-50 and ViTB/16 on the autoimmune dataset. Additional ablation studies have been provided in Appendix. 
\subsubsection{Change in Epochs}
\label{epoch-var}
In this section, we compare the performance change in CASS and DINO pretrained and then E2E finetuned with 100\% labels over the autoimmune dataset. To study the robustness, we compare the mean-variance over CNN and Transformer trained with the two techniques. The recorded mean-variance in performance for ResNet-50 and ViTB-16 trained with CASS and DINO with change in the number of pretraining epochs is 0.0001791 and 0.0002265, respectively. Based on these results, we observed that CASS-trained models have less variance, i.e., they are more robust to change in the number of pretraining epochs. 

\subsubsection{Change in Batch Size}
\label{bs-var}
Similar to Section \ref{epoch-var}, in this section, we study the change in performance concerning the batch size. As previously mentioned existing self-supervised techniques suffer a drop in performance when they are trained for small batch sizes; we studied the change in performance for batch sizes 8, 16, and 32 on the autoimmune dataset with CASS and DINO. We reported these results in Figure \ref{fig:batchsize}. We observe that the mean-variance in performance for ResNet-50 and ViTB-16 trained with CASS and DINO with change in batch size for CASS and DINO is 5.8432e-5 and 0.00015003, respectively. Hence, CASS is much more robust to changes in pretraining batch size than DINO. 

\subsubsection{Attention Maps}
\label{abl-attn-maps}
\begin{figure}[!ht]
    \centering
    \includegraphics[width=0.45\linewidth]{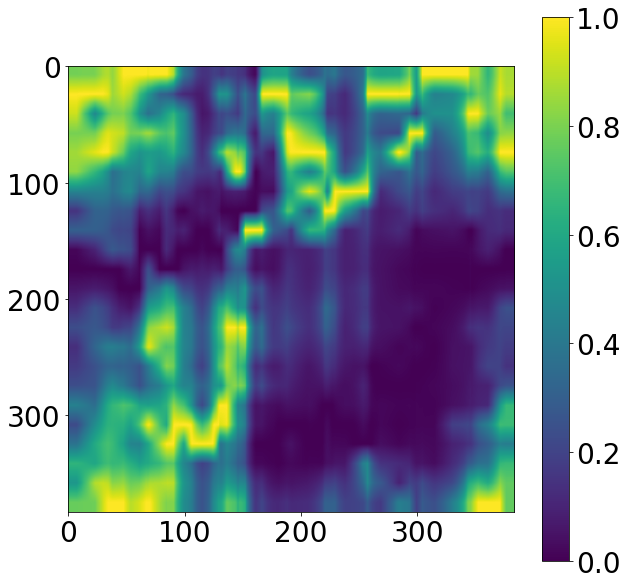}
    \includegraphics[width=0.45\linewidth]{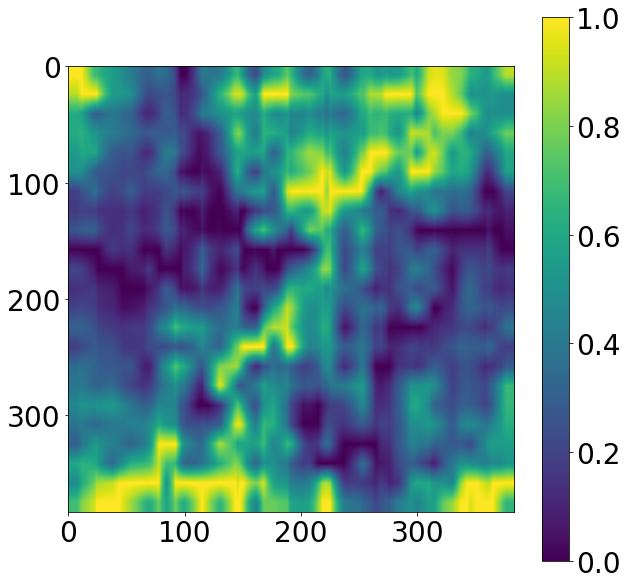}
   
    \caption{This figure shows the attention maps over a single test sample image from the autoimmune dataset. The left image is the overall attention map over a single test sample for the supervised Transformer, while the one on the right is for CASS trained Transformer.}
    \label{fig:abl_colton_avg}
\end{figure}

To study the effect qualitatively we study the attention of a supervised and CASS-pre trained Transformer. From Figure \ref{fig:abl_colton_avg} we observe that the attention map of the CASS-pre-trained Transformer is a lot more connected than a supervised Transformer due to the transfer of locality information from the CNN. We further expand on this Appendix \ref{attn-map-appendix}.  

\section{Conclusion}

Based on our experimentation on four diverse medical imaging datasets, we qualitatively concluded that CASS improves upon the classification performance of existing state-of-the-art self-supervised method DINO by 3.8\% with
1\% labeled data, 5.9\% with 10\% labeled data, and 10.13\% with 100\% labeled data and trained in 69\% less time. Furthermore, we saw that CASS is robust to batch size changes and training epochs reduction. To conclude, for medical image analysis, CASS is computationally efficient, performs better, and overcomes some of the shortcomings of existing self-supervised techniques. This ease of accessibility and better performance will catalyze medical imaging research to help us improve healthcare solutions and propagate these advancements in state-of-the-art techniques to deep practical learning in developing countries and practitioners with limited resources to develop new solutions for underrepresented and emerging diseases. 
\section*{Acknowledgements}
We would like to thank Prof. Elena Sizikova
(Moore Sloan Faculty Fellow, Center for Data Science (CDS), New York University (NYU)) for her valuable feedback and NYU HPC team for assisting us
with our computational needs.

\newpage

\bibliography{Main}
\bibliographystyle{icml2023}

\newpage
\appendix
\onecolumn

\section{CASS Pretraining Algorithm}
The core self-supervised algorithm used to train CASS with a CNN (R) and a Transformer (T) is described in Algorithm \ref{alg:algorithm}. Here, num\_epochs represents the number of self-supervised epochs to run. CNN and Transformer represent our respective architecture; for example, CNN could be a ResNet50, and Transformer can be ViT Base/16.
The loss used is described in Equation \ref{loss_eq}. Finally, after pretraining, we save the CNN and Transformer for downstream finetuning.
\begin{algorithm}[tb]
   \caption{CASS self-supervised Pretraining algorithm}
   \label{alg:algorithm}
\begin{algorithmic}
   \STATE {\bfseries Input:} Unlabeled same augmented images from the training set $x'$
   \FOR{epochs in range(num\_epochs)}
   \FOR{x in train loader:}
   \STATE $R = cnn( x')$ (taking logits output from CNN)
   \STATE $T = vit( x')$ (taking logits output from ViT)
   \IF{$R == T$}
   \STATE $R_{noise}=X \hookrightarrow  \mathcal{N}(10^{e-6}, 10^{e-9})$
   \STATE $T_{noise}=X \hookrightarrow  \mathcal{N}(10^{e-10}, 10^{e-15})$
   \STATE $R=R+R_{noise}$
   \STATE $T=T+T_{noise}$
   \STATE Calculate loss using Equation~\ref{loss_eq}
   \ELSE
   \STATE Calculate loss using Equation~\ref{loss_eq}
   \ENDIF
   
   \ENDFOR
   \ENDFOR
\end{algorithmic}
\end{algorithm}

\section{Additional Ablation Studies}
\label{aas-bs}
\subsection{Batch size}

We studied the effect of change in batch size on the autoimmune dataset in Section \ref{bs-var}. Similarly, in this section, we study the effect of varying the batch size on the brain MRI classification dataset. In the standard implementation of CASS, we used a batch size of 16; here, we showed results for batch sizes 8 and 32. The largest batch size we could run was 34 on a single GPU of 48 GB video memory. Hence 32 was the biggest batch size we showed in our results. We present these results in Table \ref{batchsize_bmri}. Similar to the results in Section \ref{bs-var}, performance decreases as we reduce the batch size and increases slightly as we increase the batch size for both CNN and Transformer. 

\begin{table}[!htb]
\centering
\begin{tabular}{lll}
\hline
Batch Size & CNN F1 Score                       & Transformer F1 Score \\
\hline
8          & 0.9895±0.0025                      & 0.9198±0.0109       \\
16         & \multicolumn{1}{c}{0.9909± 0.0032} & 0.9279± 0.0213       \\
32         & 0.991±0.011                       & 0.9316±0.006        \\
\hline
\end{tabular}
\caption{This table represents the results for different batch sizes on the brain MRI classification dataset.  We maintain the downstream batch size constant in all three cases, following the standard experimental setup mentioned in Appendix \ref{ssl-train} and \ref{sup-train}. These results are on the test set after E2E fine-tuning with 100\% labels.}
\label{batchsize_bmri}
\end{table}

\subsection{Change in pretraining epochs}

As standard, we pretrained CASS for 100 epochs in all cases. However, existing self-supervised techniques are plagued with a loss in performance with a decrease in the number of pretraining epochs. To study this effect for CASS, we reported results in Section \ref{epoch-var}. Additionally, in this section, we report results for training CASS for 300 epochs on the autoimmune and brain tumor MRI datasets. We reported these results in Table \ref{epochs_atm} and \ref{epochs_bmri}, respectively. We observed a slight gain in performance when we increased the epochs from 100 to 200 but minimal gain beyond that. We also studied the effect of longer pretraining on the  brain tumor MRI classification dataset and presented these results in Table \ref{epochs_bmri}. 
\begin{table}[!htb]
\centering
\begin{tabular}{lll}
\hline
\multicolumn{1}{c}{Epochs} & CNN F1 Score  & Transformer F1 Score \\
\hline
50                         & 0.8521±0.0007 & 0.8765± 0.0021       \\
100                        & 0.8650±0.0001 & 0.8894±0.005         \\
200                        & 0.8766±0.001  & 0.9053±0.008         \\
300                        & 0.8777±0.004  & 0.9091±8.2e-5       \\
\hline
\end{tabular}
\caption{ Performance comparison over a varied number of epochs on the brain tumor MRI classification dataset, from 50 to 300 epochs, the downstream training procedure, and the CNN-Transformer combination is kept constant across all the four experiments, only the number of self-supervised pretraining epochs were changed.}
\label{epochs_atm}
\end{table}

\begin{table}[!htb]
\centering
\begin{tabular}{lll}
\hline
Epochs & CNN F1 Score                       & Transformer F1 Score \\
\hline
50                              & 0.9795±0.0109                      & 0.9262±0.0181       \\
100                             & \multicolumn{1}{c}{0.9909± 0.0032} & 0.9279± 0.0213       \\
200                             & 0.9864±0.008                       & 0.9476±0.0012        \\
300                             & 0.9920±0.001                       & 0.9484±0.017   \\
\hline
\end{tabular}
\caption{ Performance comparison over a varied number of epochs, from 50 to 300 epochs, the downstream training procedure, and the CNN-transformer combination is kept constant across all four experiments; only the number of self-supervised epochs has been changed.}
\label{epochs_bmri}
\end{table}

\subsection{Augmentations}

Contrastive learning techniques are known to be highly dependent on augmentations. Recently, most self-supervised techniques have adopted BYOL \cite{Grill2020BootstrapYO}-like a set of augmentations. DINO \cite{caron2021emerging} uses the same set of augmentations as BYOL, along with adding local-global cropping. We use a reduced set of BYOL augmentations for CASS, along with a few changes. For instance, we do not use solarize and Gaussian blur. Instead, we use affine transformations and random perspectives. In this section, we study the effect of adding BYOL-like augmentations to CASS. We report these results in Table \ref{augmentaions}. We observed that CASS-trained CNN is robust to changes in augmentations. On the other hand, the Transformer drops performance with changes in augmentations. A possible solution to regain this loss in performance for Transformer with a change in augmentation is using Gaussian blur, which converges the results of CNN and the Transformer. 

\begin{table*}[!htb]
\label{table:augchnages}
\centering
\begin{tabular}{cll}
\hline
Augmentation Set                                    & CNN F1 Score    & Transformer F1 Score \\
\hline
CASS only                                           & 0.8650±0.0001   & 0.8894±0.005         \\
CASS + Solarize                                     & 0.8551±0.0004   & 0.81455±0.002        \\
CASS + Gaussian blur                                & 0.864±4.2e-05   & 0.8604±0.0029        \\

\multicolumn{1}{l}{CASS + Gaussian blur + Solarize} & 0.8573±2.59e-05 & 0.8513±0.0066       \\
\hline
\end{tabular}
\caption{We report the F1 metric of CASS trained with
a different set of augmentations for 100 epochs. While CASS-trained CNN fluctuates within a percent of its peak performance, CASS-trained Transformer drops performance with the addition of solarization and Gaussian blur. Interestingly, the two arms converged with the use of Gaussian blur.}
\label{augmentaions}
\end{table*}

\subsection{Optimization}
\label{optim-results}
In CASS, we use Adam optimizer for both CNN and Transformer. This is a shift from using SGD or stochastic gradient descent for CNNs. In this Table \ref{optimizer}, we report the performance of CASS-trained CNN and Transformer with the CNN using SGD and Adam optimizer. We observed that while the performance of CNN remained almost constant, the performance of the Transformer dropped by almost 6\% with CNN using SGD.

\begin{table}[!htb]
\centering
\begin{tabular}{cll}
\hline
Optimiser for CNN                                   & CNN F1 Score    & Transformer F1 Score \\
\hline
Adam                                                & 0.8650±0.0001   & 0.8894±0.005         \\
SGD                                                 & 0.8648±0.0005   & 0.82355±0.0064       \\
\hline
\end{tabular}
\caption{We report the F1 metric of CASS trained with
a different set of optimizers for the CNN arm for 100 epochs. While there is no change in CNN's performance, the Transformer's performance drops around 6\% with SGD.}
\label{optimizer}
\end{table}

\subsection{Using softmax and sigmoid layer in CASS}

As noted in Fig \ref{fig:CASS}, CASS doesn’t use a softmax layer like DINO \cite{caron2021emerging} before the computing loss. The output logits of the two networks have been used to combine the two architectures in a response-based knowledge distillation \cite{gou2021knowledge} manner instead of using soft labels from the softmax layer. In this section, we study the effect of using an additional softmax layer on CASS. Furthermore, we also study the effect of adding a sigmoid layer instead of a softmax layer and compare it with a CASS model that doesn’t use the sigmoid or the softmax layer. We present these results in Table \ref{softmax}. We observed that not using sigmoid and softmax layers in CASS yields the best result for both CNN and Transformers.

\begin{table}[!htb]
\centering
\begin{tabular}{lll}
\hline
Techniques & CNN F1 Score  & Transformer F1 Score \\
\hline
Without Sigmoid or Softmax       & 0.8650±0.0001  & 	0.8894±0.005         \\
With Sigmoid Layer      & 0.8296±0.00024 & 0.8322±0.004      \\
With Softmax Layer      & 0.8188±0.0001 & 0.8093±0.00011      \\
\hline
\end{tabular}
\caption{We observe that performance reduces when we introduce the sigmoid or softmax layer.}
\label{softmax}
\end{table}

\subsection{Change in architecture}
\label{change-arch}

\subsubsection{Changing Transformer and keeping the CNN same}

From Table \ref{differentTrasnformer} and \ref{differentTrasnformer_results}, we observed that CASS-trained ViT Transformer with the same CNN consistently gained approximately 4.7\% over its
supervised counterpart. Furthermore, from Table \ref{differentTrasnformer_results}, we observed that although ViT L/16 performs better than ViT B/16 on ImageNet ( \cite{rw2019timm}'s results), we observed that the trend is opposite on the autoimmune dataset. Hence, the supervised performance of architecture must be considered before pairing it with CASS.

\begin{table}[!htb]
\centering
\begin{tabular}{lll}
\hline
 Transformer            & CNN F1 Score  & Transformer F1 Score \\
\hline
 ViT Base/16   & 0.8650±0.001 & 0.8894± 0.005       \\
ViT Large/16  & 0.8481±0.001 & 0.853±0.004       \\
\hline
\end{tabular}
\caption{In this table, we show the performance of CASS for ViT large/16 with ResNet-50 and ViT base/16 with ResNet-50. We observed that CASS-trained Transformers, on average, performed 4.7\% better than their supervised counterparts.}
\label{differentTrasnformer}
\end{table}

\begin{table}[!htb]
\centering
\begin{tabular}{ll}
\hline
\multicolumn{1}{c}{Architecture} & Testing F1 Score \\
\hline
ResnNet-50                       & 0.83895±0.007     \\
ViT Base/16                      & 0.8420±0.009     \\
ViT large/16                     & 0.80495±0.0077 \\
\hline
\end{tabular}
\caption{Supervised performance of ViT family on the autoimmune dataset. We observed that as opposed to ImageNet performance, ViT large/16 performs worse than ViT Base/16 on the autoimmune dataset.}
\label{differentTrasnformer_results}
\end{table}

We keep the CNN constant for this experiment and study the effect of changing the Transformer. For this experiment, we use ResNet as our choice of CNN and ViT base and large Transformers with 16 patches. Additionally, we also report performance for DeiT-B \cite{touvron2020training} with ResNet-50. We report these results in Table \ref{transformer_bmri}. Similar to Table \ref{differentTrasnformer}, we observe that by changing Transformer from ViT Base to Large while keeping the number of tokens the same at 16, performance drops. Additionally, for approximately the same size, out of DeiT base and ViT base Transformers, DeiT performs much better than ViT base. 

\begin{table*}[t]
\centering
\begin{tabular}{clll}
\hline
CNN                                                                           & Transformer            & CNN F1 Score  & Transformer F1 Score \\
\hline
\multirow{3}{*}{\begin{tabular}[c]{@{}c@{}}ResNet-50\\ (25.56M)\end{tabular}} & DEiT Base/16 (86.86M)     & 0.9902±0.0025   & 0.9844±0.0048     \\
& ViT Base/16 (86.86M)   & 0.9909±0.0032 & 0.9279± 0.0213       \\
                                                                              & ViT Large/16 (304.72M) & 0.98945±2.45e-5 & 0.8896±0.0009       \\
                                                                              
\hline
\end{tabular}
\caption{For the same number of Transformer parameters, DEiT-base with ResNet-50 performed much better than ResNet-50 with ViT-base. The difference in their CNN arm is ~0.10\%. On ImageNet DEiT-base has a top1\% accuracy of 83.106 while ViT-base has an accuracy of 86.006. We use both Transformers with 16 patches. [ResNet-50 has an accuracy of 80.374] }
\label{transformer_bmri}
\end{table*}

\subsubsection{Changing CNN and keeping the Transformer same}

Table \ref{sameTransformer_atm} and \ref{cnn_atm} we observed that similar to changing Transformer while keeping the CNN same, CASS-trained CNNs gained an average of 3\% over their supervised counterparts. ResNet-200 \cite{rw2019timm} doesn't have ImageNet initialization hence using random initialization. 

\begin{table*}[!htb]
\centering
\begin{tabular}{llll}
\hline
CNN                           & Transformer                  & \multicolumn{2}{l}{100\% Label Fraction} \\
                              &                              & CNN F1 score     & Transformer F1 score   \\
\hline
\multicolumn{1}{c}{ResNet-18 (\textbf{11.69M})} & \multirow{3}{*}{ViT Base/16 (\textbf{86.86M})} & 0.8674±4.8e-5     & 0.8773±5.29e-5           \\
ResNet-50 (\textbf{25.56M})                     &                              & 0.8680±0.001   & 0.8894± 0.0005         \\
ResNet-200 (\textbf{64.69M})                    &                              & 0.8517±0.0009     & 0.874±0.0006     \\
\hline

\end{tabular}
\caption{F1 metric comparison between the two arms of CASS trained over 100 epochs, following the protocols and procedure listed in Appendix E. The numbers in parentheses show the parameters learned by the network. We use \cite{rw2019timm} implementation of CNN and transformers, with ImageNet initialization except for ResNet-200.}
\label{sameTransformer_atm}
\end{table*}

\begin{table}[!htb]
\centering
\begin{tabular}{ll}
\hline
\multicolumn{1}{c}{Architecture} & Testing F1 Score                                        \\
\hline
ResnNet-18                       & 0.8499±0.0004                                           \\
ResnNet-50                       & 0.83895±0.007                                            \\
ResnNet-200                      & 0.833±0.0005  \\
\hline
\end{tabular}
\caption{Supervised performance of the ResNet CNN family on the autoimmune dataset.}
\label{cnn_atm}
\end{table}

For this experiment, we use the ResNet family of CNNs and ViT base/16 as our Transformer. We use ImageNet initialization for ResNet 18 and 50, while random initialization for ResNet-200 (As Timm's library doesn't have an ImageNet initialization). We present these results in Table \ref{differentcnn_bmri}. We observed that an increase in the performance of ResNet correlates to an increase in the performance of the Transformer, implying that there is information transfer between the two.

\begin{table*}[!htb]
\centering
\begin{tabular}{llll}
\hline
CNN                           & Transformer                  & \multicolumn{2}{l}{100\% Label Fraction} \\
                              &                              & CNN F1 score     & Transformer F1 score   \\
\hline
\multicolumn{1}{c}{ResNet-18 (\textbf{11.69M})} & \multirow{3}{*}{ViT Base/16 (\textbf{86.86M})} & 0.9913±0.002     & 0.9801±0.007           \\
ResNet-50 (\textbf{25.56M})                     &                              & 0.9909±0.0032    & 0.9279± 0.0213         \\
ResNet-200 (\textbf{64.69M})                    &                              & 0.9898±0.005     & 0.9276±0.017     \\
\hline

\end{tabular}
\caption{F1 metric comparison between the two arms of CASS trained over 100 epochs, following the protocols and procedure listed in Appendix \ref{ssl-train} and \ref{sup-train}. The numbers in parentheses show the parameters learned by the network. We use \cite{rw2019timm} implementation of CNN and transformers, with ImageNet initialization except for ResNet-200.}
\label{differentcnn_bmri}
\end{table*}

\subsubsection{Using CNN in both arms}

We have experimented using a CNN and a Transformer in CASS on the brain tumor MRI classification dataset. In this section, we present results for using two CNNs in CASS. We pair ResNet-50 with DenseNet-161. We observe that both CNNs fail to reach the benchmark set by ResNet-50 and ViT-B/16 combination. Although training the ResNet-50-DenseNet-161 pair takes 5 hours 24 minutes, less than the 7 hours 11 minutes taken by the ResNet-50-ViT-B/16 combination to be trained with CASS. We compare these results in Table \ref{bothcnn}.

\begin{table*}[!htb]
\centering
\begin{tabular}{clll}
\hline
CNN                        & \begin{tabular}[c]{@{}l@{}}Architecture in\\  arm 2\end{tabular} & F1 Score of ResNet-50 arm  & F1 Score of arm 2 \\
\hline
\multirow{2}{*}{ResNet-50} & ViT Base/16                                                              & 0.9909±0.0032 & 0.9279± 0.0213            \\
                           & DenseNet-161                                                             & 0.9743±8.8e-5 & 0.98365±9.63e-5    \\
\hline
\end{tabular}
\caption{We observed that for the ResNet-50-DenseNet-161 pair, we train 2 CNNs instead of 1 in our standard setup of CASS. Furthermore, none of these CNNs could match the performance of ResNet-50 trained with the ResNet-50-ViT base/16 combination. Hence, by adding a Transformer-CNN combination, we transfer information between the two architectures that would have been missed otherwise.}
\label{bothcnn}
\end{table*}

\subsubsection{Using Transformer in both arms}

Similar to the above section, we use a Transformer-Transformer combination instead of a CNN-Transformer combination. We use Swin-Transformer patch-4/window-12 \cite{Liu_2021_ICCV} alongside ViT-B/16 Transformer. We observe that the performance for ViT/B-16 improves by around ~1.3\% when we use Swin Transformer. However, this comes at a computational cost. The swin-ViT combination took 10 hours to train as opposed to 7 hours and 11 minutes taken by the ResNet-50-ViT-B/16 combination to be trained with CASS. Even with the increased time to train the Swin-ViT combination, it is still almost 50\% less than DINO.
We present these results in Table \ref{bothT}.

\begin{table*}[!htb]
\centering
\begin{tabular}{clll}
\hline
\begin{tabular}[c]{@{}c@{}}Architecture in\\  arm 1\end{tabular} & Transformer                   & F1 Score of arm 1  & F1 Score of ViT-B/16 arm \\
\hline
\multicolumn{1}{l}{ResNet-50}                                            & \multirow{2}{*}{ViT Base/16} & 0.9909±0.0032  & 0.9279± 0.0213            \\
Swin Transformer                                                          &                              & 0.9883±1.26e-5 & 0.94±8.12e-5     \\
\hline
\end{tabular}
\caption{We present the results for using Transformers in both arms and compare the results with the CNN-Transformer combination.}
\label{bothT}
\end{table*}

\subsection{Effect of Initialization}
Although the aim of self-supervised pretraining is to provide better initialization, we use ImageNet initialized CNN and Transformers for CASS and DINO pertaining as well as supervised training similar to \cite{Matsoukas2021IsIT}. We use Timm's library for these initialization \cite{rw2019timm}. ImageNet initialization is preferred not because of feature reuse but because ImageNet weights allow for faster convergence through better weight scaling \cite{raghu2019transfusion}. But sometimes pre-trained weights might be hard to find, so we study CASS' performance with random and ImageNet initialization in this section.
We observed that performance almost remained the same, with minor gains when the initialization was altered for the two networks. Table \ref{init} presents the results of this experimentation.

\begin{table}[!htb]
\centering
\begin{tabular}{lll}
\hline
Initialisation & CNN F1 Score  & Transformer F1 Score \\
\hline
Random        & 0.9907±0.009  & 0.9116±0.027         \\
Imagenet      & 0.9909±0.0032 & 0.9279± 0.0213      \\
\hline
\end{tabular}
\caption{We observe that the Transformer gains some performance with the random initialization, although performance has more variance when used with random initialization.}
\label{init}
\end{table}

\begin{table}[!htb]
\centering
\begin{tabular}{lll}
\hline
Initialisation & CNN F1 Score  & Transformer F1 Score \\
\hline
Random        & 0.8437±0.0047  & 0.8815±0.048        \\
Imagenet      & 0.8650±0.0001 & 0.8894±0.005      \\
\hline
\end{tabular}
\caption{We observe that the Transformer gains some performance with the random initialization, although performance has more variance when used with random initialization.}
\label{init-ai}
\end{table}

\section{Result Analysis}

\subsection{Time complexity analysis}
In Section \ref{time}, we observed that CASS takes 69\% less time than DINO. This reduction in time could be attributed to the following reasons:
\begin{enumerate}
    \item In DINO, augmentations are applied twice as opposed to just once in CASS. Furthermore, per application, CASS uses fewer augmentations than DINO.
    \item Since the architectures used are different, there is no scope for parameter sharing between them. A major chunk of time is saved by updating the two architectures after each epoch instead of re-initializing architectures with lagging parameters.  
\end{enumerate}
\subsection{Qualitative analysis}
\label{Qual}
To qualitatively expand our study, in this section, we study the feature maps of CNN and attention maps of Transformers trained using CASS and supervised techniques. To reinstate, based on the study by \cite{Raghu2021DoVT}, since CNN and Transformer extract different kinds of features from the same input, combing the two of them would help us create positive pairs for self-supervised learning. In doing so, we would transfer between the two architectures, which is not innate. We have already seen that this yield better performance in most cases over four different datasets and with three different label fractions. In this section, we study this gain qualitatively with the help of feature maps and class attention maps. Also, we briefly discussed attention maps in Section \ref{abl-attn-maps}, where we observed that CASS-trained Transformers have more local understanding of the image and hence a more connected attention map than purely-supervised Transformer.

\subsection{Feature maps}
In this section, we study the feature maps from the first five layers of the ResNet-50 model trained with CASS and supervision. We extracted feature maps after the Conv2d layer of ResNet-50. We present the extracted features in Figure \ref{fig:fmaps}. We observed that CASS-trained CNN could retain much more detail about the input image than supervised CNN.

\begin{figure}
    \centering
    \includegraphics[width=0.5\linewidth]{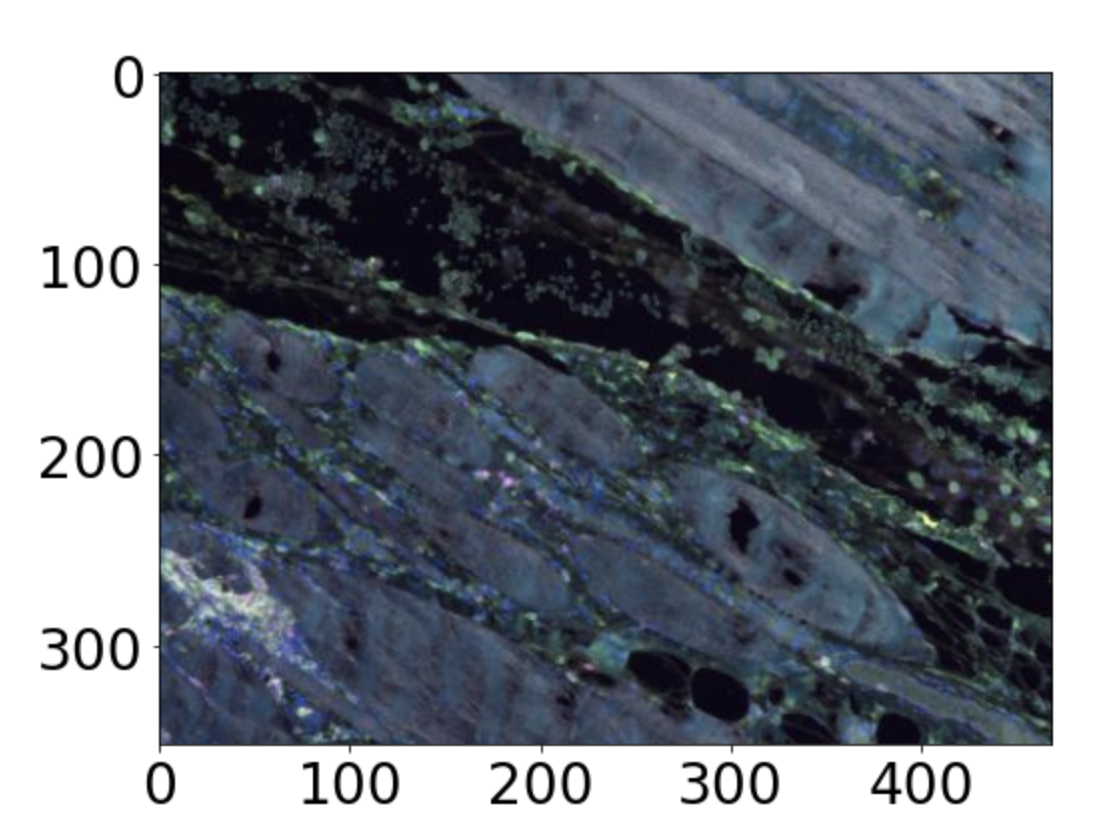}
    \caption{Sample image used from the test set of the autoimmune dataset.}
    \label{fig:sample_image}
\end{figure}

\begin{figure}[!ht]
    \centering
    \includegraphics[width=1\linewidth]{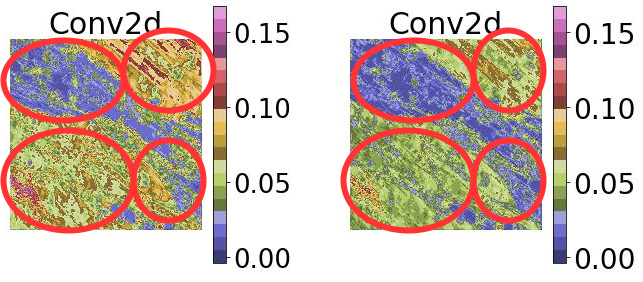}
    \caption{This figure shows the feature map extracted after the first Conv2d layer of ResNet-50 for CASS (on the left) and supervised CNN (on the right). The color bar shows the intensity of pixels retained. From the four circles, it is clear that CASS-trained CNN can retain more intricate details about the input image (Figure \ref{fig:sample_image}) more intensely so that they can be propagated through the architecture and help the model learn better representations as compared to the supervised CNN. We study the same feature map in detail for the first five layers after Conv2d in Figure \ref{fig:fmaps}.}
    
    \label{fig:fmapssingle}
\end{figure}

\begin{figure*}[t]
    \includegraphics[width=0.9\linewidth]{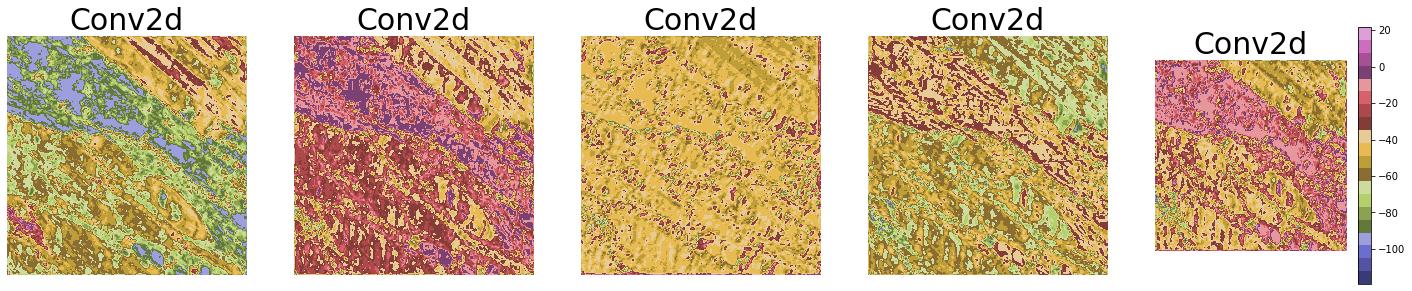}
    \includegraphics[width=0.9\linewidth]{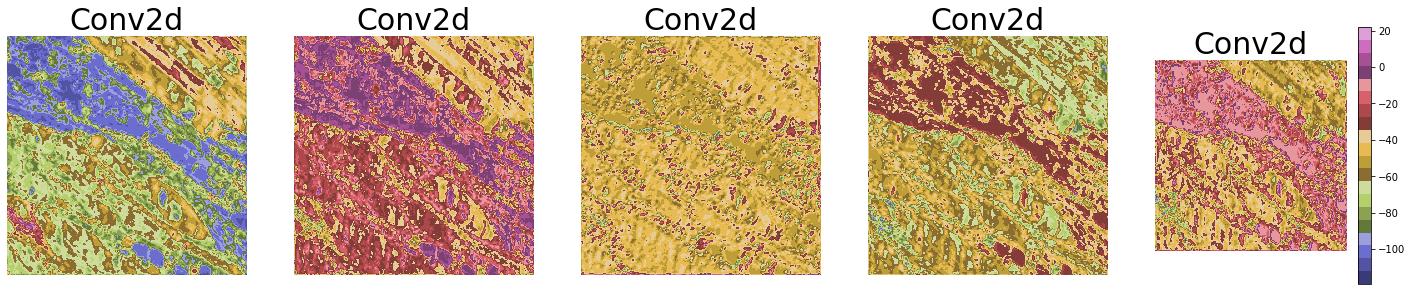}
    \caption{ At the top, we have features extracted from the top 5 layers of supervised ResNet-50, while at the bottom, we have features extracted from the top 5 layers of CASS-trained ResNet-50. We supplied both networks with the same input ( shown in Figure~\ref{fig:sample_image}). }
    \label{fig:fmaps}
\end{figure*}

\subsection{Class attention maps}
\label{attn-map-appendix}

We have already studied the class attention maps over a single image in Section \ref{abl-attn-maps}. This section will explore the average class attention maps for all four datasets. We studied the attention maps averaged over 30 random samples for autoimmune, dermofit, and brain MRI datasets. Since the ISIC 2019 dataset is highly unbalanced, we averaged the attention maps over 100 samples so that each class may have an example in our sample. We maintained the same distribution as the test set, which has the same class distribution as the overall training set. We observed that CASS-trained Transformers were better able to map global and local connections due to Transformers' ability to map global dependencies and by learning features sensitive to translation equivariance and locality from CNN. This helps the Transformer learn features and local patterns that it would have missed.

\subsubsection{Autoimmune dataset}
\begin{figure}[!ht]
    \centering
    \includegraphics[width=0.4\linewidth]{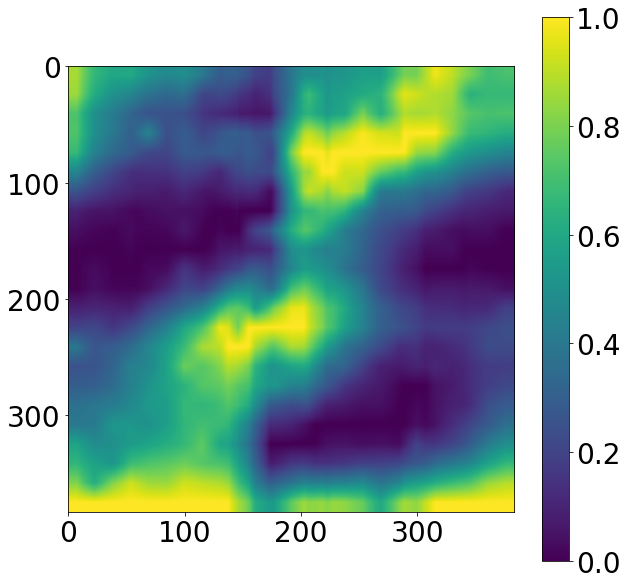}
    \includegraphics[width=0.4\linewidth]{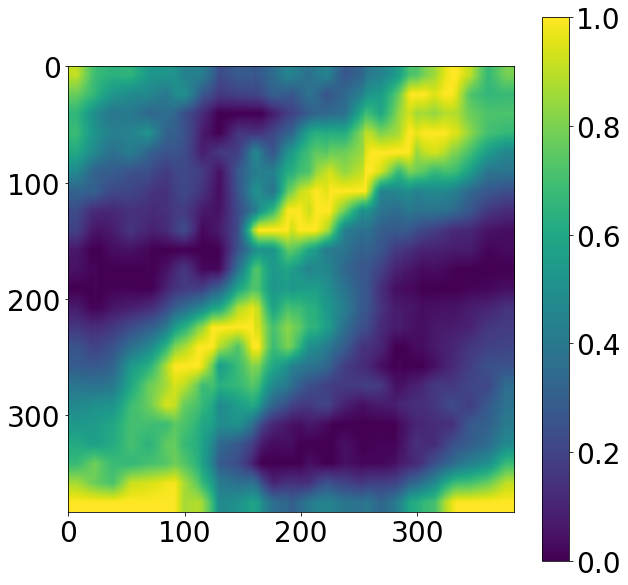}
   
    \caption{To ensure the consistency of our study, we studied average attention maps over 30 sample images from the autoimmune dataset. The left image is the overall attention map averaged over 30 samples for the supervised Transformer, while the one on the right is for CASS pretrained Transformer (both after finetuning with 100\% labels).}
    \label{fig:colton_avg}
\end{figure}

We study the class attention maps averaged over 30 test samples for the autoimmune dataset in Figure \ref{fig:colton_avg}. We observed that the CASS-trained Transformer has much more attention in the center than the supervised Transformer. This extra attention could be attributed to a Transformer on its own inability to map out due to the information transfer from CNN. Another feature to observe is that the attention map of the CASS-trained Transformer is much more connected than that of a supervised Transformer.

    \begin{figure}[!hb]
    \centering
    \includegraphics[width=0.4\linewidth]{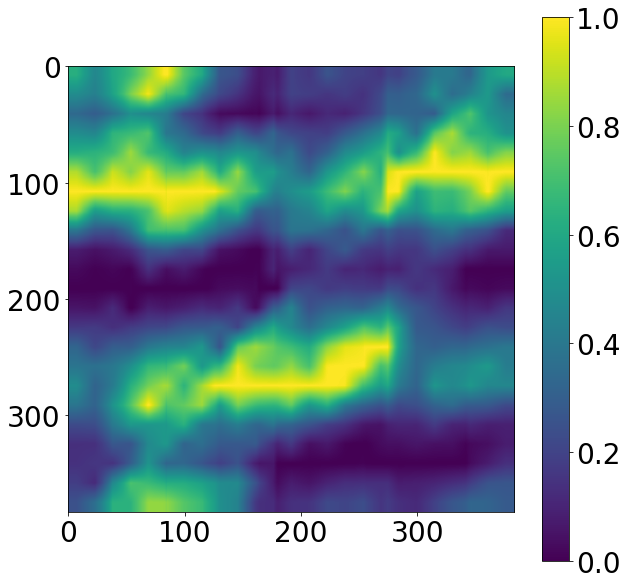}
    \includegraphics[width=0.4\linewidth]{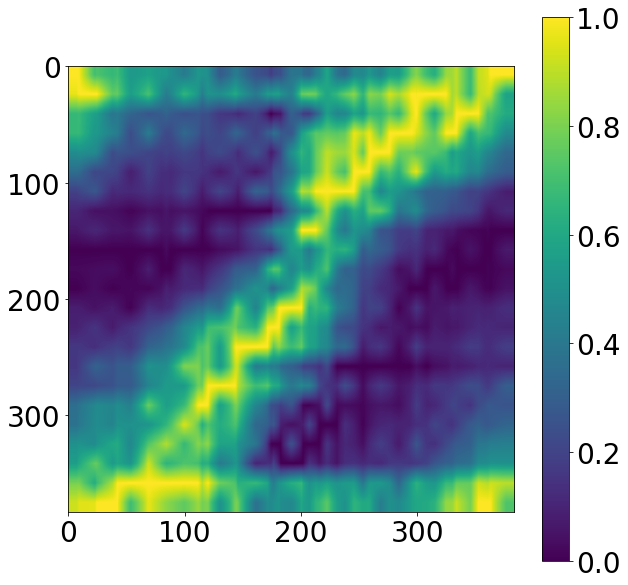}
    
    \caption{Class attention maps averaged over 30 samples of the dermofit dataset for supervised Transformer (on the left), and CASS pretrained Transformer (on the right). Both after finetuning with 100\% labels.}
    
    \label{fig:dermofit_attn}
\end{figure}

\subsubsection{Dermofit dataset} We present the average attention maps for the dermofit dataset in Figure \ref{fig:dermofit_attn}. We observed that the CASS-trained Transformer can pay much more attention to the center part of the image. Furthermore, the attention map of the CASS-trained Transformer is much more connected than the supervised Transformer. So, overall with CASS, the Transformer is not only able to map long-range dependencies which are innate to Transformers but is also able to make more local connections with the help of features sensitive to translation equivariance and locality from CNN.
\begin{figure}[!ht]
    \centering
    \includegraphics[width=0.4\linewidth]{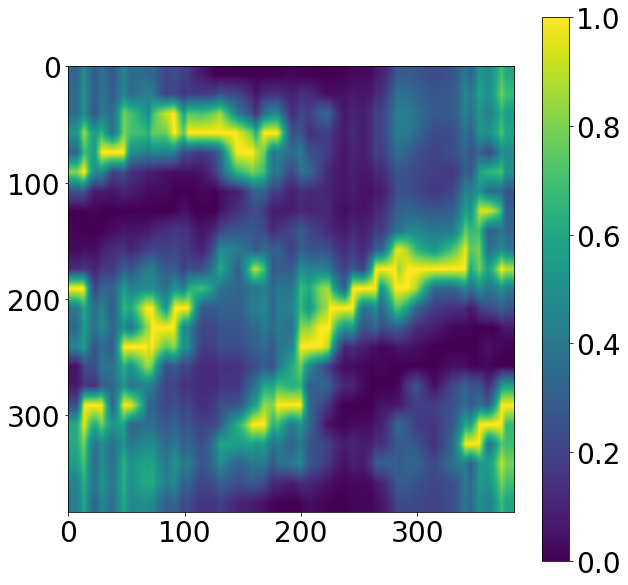}
    \includegraphics[width=0.4\linewidth]{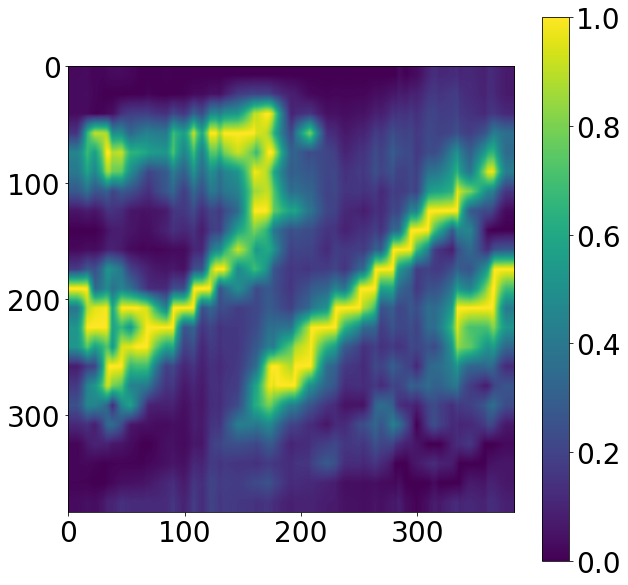}
    \caption{Class attention maps averaged over 30 samples of the brain tumor MRI classification dataset for supervised Transformer (on the left), and CASS pretrained Transformer (on the right). Both after finetuning with 100\% labels.}
    
    \label{fig:bmri_attn}
\end{figure}
    

\subsubsection{Brain tumor MRI classification dataset} We present the average class attention maps results in Figure \ref{fig:bmri_attn}. We observed that a CASS-trained Transformer could better capture long and short-range dependencies than a supervised Transformer. Furthermore, we observed that a CASS-trained Transformer's attention map is much more centered than a supervised Transformer's. From Figure \ref{fig:bmri}, we can observe that most MRI images are center localized, so having a more centered attention map is advantageous in this case.   

\subsubsection{ISIC 2019 dataset} 
\begin{figure}[!ht]
    \centering
    \includegraphics[width=0.4\linewidth]{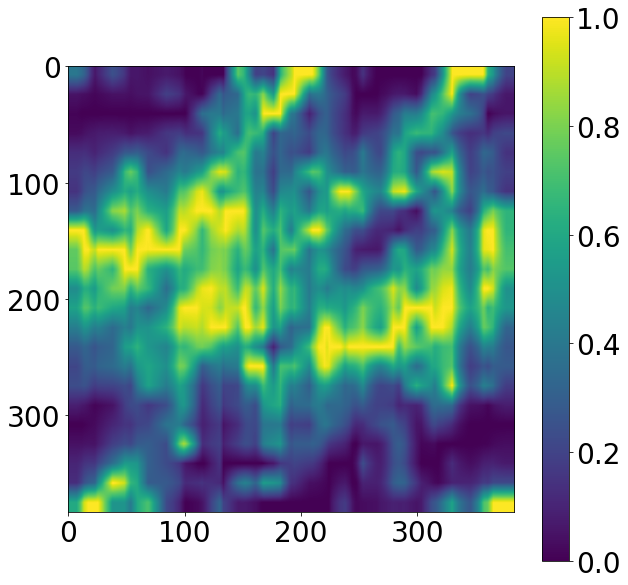}
    \includegraphics[width=0.4\linewidth]{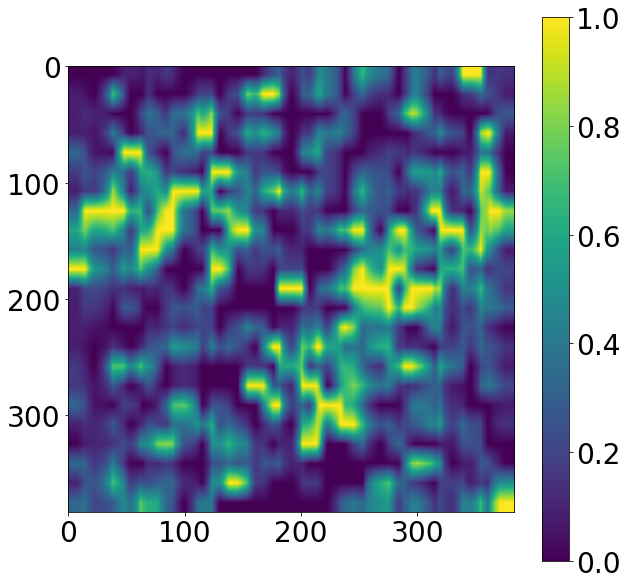}
    \caption{Class attention maps averaged over 100 samples from the ISIC-2019 dataset for the supervised Transformer (on the left) and CASS-trained Transformer (on the right). Both after finetuning with 100\% labels.}
    
    \label{fig:isic_attn}
\end{figure}

The ISIC-2019 dataset is one of the most challenging datasets out of the four datasets. ISIC 2019 consists of images from the HAM10000 and BCN\_20000 datasets \cite{cassidy2022analysis, Gessert2020SkinLC}. For the HAM1000 dataset, it isn't easy to classify between 4 classes (melanoma and melanocytic nevus), (actinic keratosis, and benign keratosis). HAM10000 dataset contains images of size 600×450, centered and cropped around the lesion. Histogram corrections have been applied to only a few images. The BCN\_20000 dataset contains images of size 1024×1024. This dataset is particularly challenging as many images are uncropped, and lesions are in difficult and uncommon locations. Hence, in this case, having more spread-out attention maps would be advantageous instead of a more centered one. From Figure \ref{fig:isic_attn}, we observed that a CASS-trained Transformer has a lot more spread attention map than a supervised Transformer. Furthermore, a CASS-trained Transformer can also attend the corners far better than a supervised Transformer.  

From Figures \ref{fig:colton_avg}, \ref{fig:dermofit_attn}, \ref{fig:bmri_attn} and \ref{fig:isic_attn}, we observed that in most cases, the supervised Transformer had spread out attention, while the CASS trained Transformer has a more "connected."
attention map. This is primarily because of local-level information transfer from CNN. Hence we could add some more image-level intuition, with the help of CNN, to the Transformer that it would have rather missed on its own.

\subsubsection{Choice of Datasets}
\label{ds-choice}
\begin{figure}[!h]
    \centering
    \includegraphics[width=0.4\linewidth]{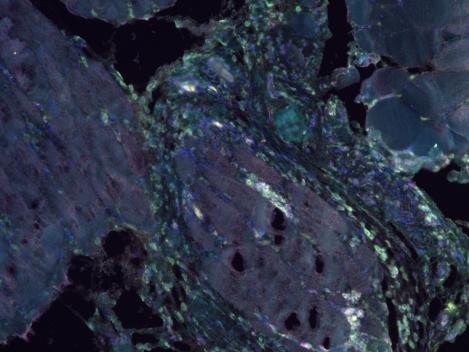}
    \includegraphics[width=0.4\linewidth]{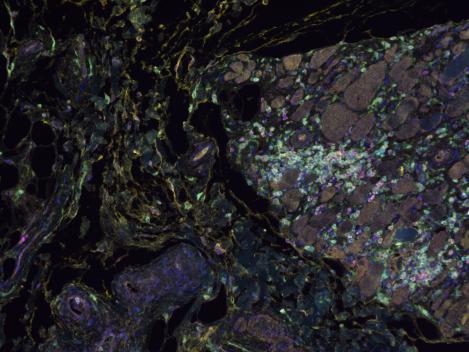}
    \caption{Sample of autofluorescence slide images from the muscle biopsy of patients with dermatomyositis - a type of autoimmune disease.}
    \label{fig:autoimmune}
\end{figure}

\begin{figure}[!htb]
    \centering  
    \includegraphics[width=0.3\linewidth]{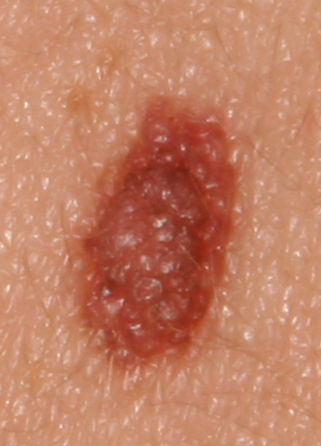}
    \includegraphics[width=0.3\linewidth]{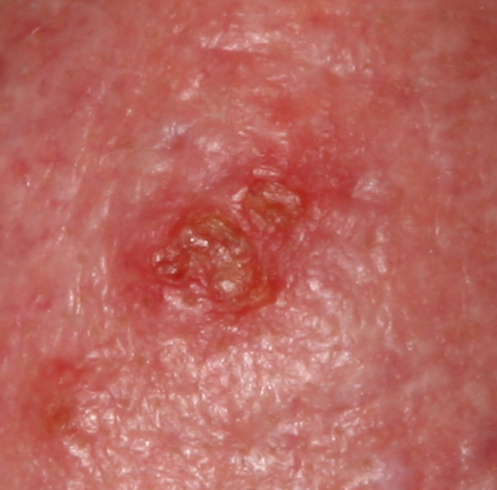}
    \includegraphics[width=0.3\linewidth]{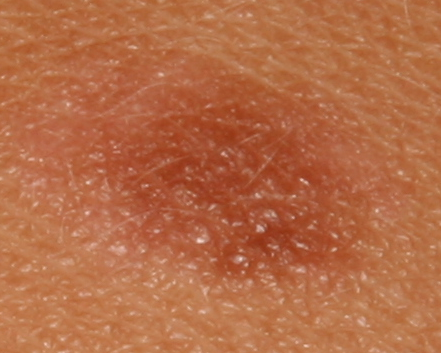}
     \hspace{1cm}
    \caption{Sample images from the Dermofit dataset.}
    \label{fig:dermofit}
    \end{figure}

\begin{figure}[!htb]
     \centering
    \includegraphics[width=0.25\linewidth]{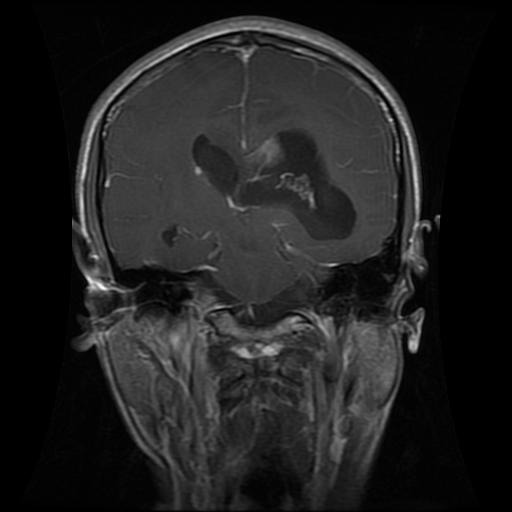}
    \includegraphics[width=0.25\linewidth]{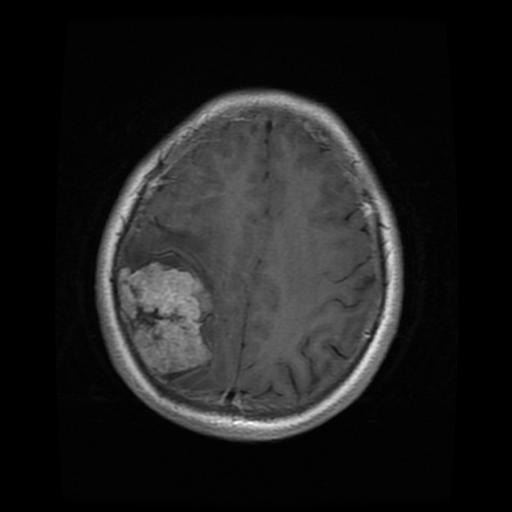}
    \includegraphics[width=0.25\linewidth]{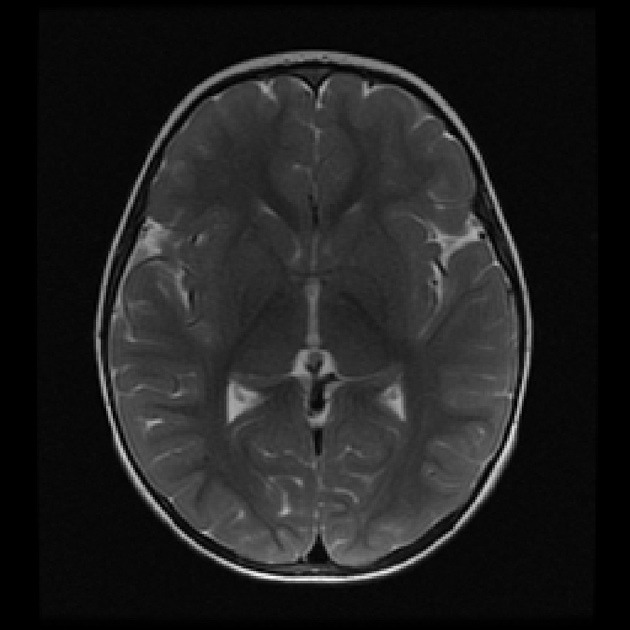}
     \hspace{1cm}
    \caption{Sample images of brain tumor MRI dataset, Each image corresponds to a prediction class in the data set glioma (Left), meningioma (Center), and No tumor (Right)  }
    \label{fig:bmri}
    \end{figure}

\begin{figure}[!htb]
    \centering  
    \includegraphics[width=0.25\linewidth]{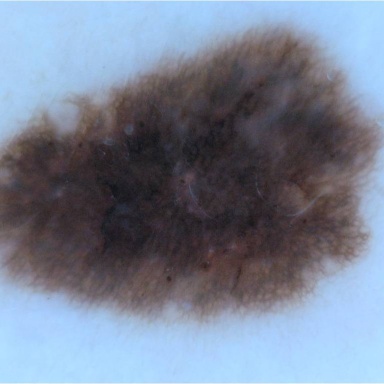}
    \includegraphics[width=0.25\linewidth]{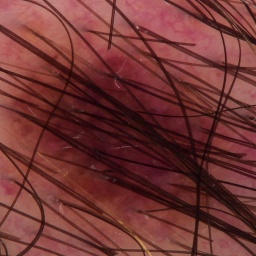}
    \includegraphics[width=0.25\linewidth]{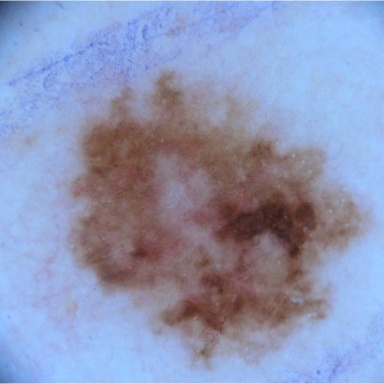}
     \hspace{1cm}
    \caption{Sample images from the ISIC-2019 challenge dataset.}
    \label{fig:isic}
    \end{figure}

We chose four medical imaging datasets with diverse sample sizes ranging from 198 to 25,336 and diverse modalities to study the performance of existing self-supervised techniques and CASS.
Most of the existing self-supervised techniques have been studied on million image datasets, but medical imaging datasets, on average, are much smaller than a million images. Furthermore, they are usually imbalanced and some of the existing self-supervised techniques rely on batch statistics, which makes them learn skewed representations. We also include a dataset of emerging and underrepresented diseases with only a few hundred samples, the autoimmune dataset in our case (198 samples). To the best of our knowledge, no existing literature studies the effect of self-supervised learning on such a small dataset. Furthermore, we chose the dermofit dataset because all the images are taken using an SLR camera, and no two images are the same size. Image size in dermofit varies from 205×205 to 1020×1020. MRI images constitute a large part of medical imaging; hence we included this dataset in our study. So, to incorporate them in our study, we had the Brain tumor MRI classification dataset. Furthermore, it is our study's only black-and-white dataset; the other three datasets are RGB.
The ISIC 2019 is a unique dataset as it contains multiple pairs of hard-to-classify classes (Melanoma - melanocytic nevus and actinic keratosis - benign keratosis) and different image sizes - out of which only a few have been prepossessed. It is a highly imbalanced dataset containing samples with lesions in difficult and uncommon locations. To give an idea about the images used in our experiments, we provide sample images from the four datasets used in our experimentation in Figures \ref{fig:autoimmune}, \ref{fig:dermofit}, \ref{fig:bmri} and \ref{fig:isic}. 

\subsection{Self-supervised pretraining}
\label{ssl-train}
\subsubsection{Protocols}
\begin{itemize}
    \item Self-supervised learning was only done on the training data and not on the validation data. We used \url{https://github.com/PyTorchLightning/pytorch-lightning} to set the pseudo-random number generators in PyTorch, NumPy, and (python.random).

   \item We ran training over five seed values and reported mean results with variance in each table. We didn't perform a seed value sweep to extract any more performance \cite{picard2021torch}.
   
   \item For DINO implementation, we use Phil Wang's implementation: \url{https://github.com/lucidrains/vit-pytorch}.
   
   \item For the implementation of CNNs and Transformers, we use Timm's library \cite{rw2019timm}.
   
   \item For all experiments, ImageNet~\cite{deng2009imagenet} initialised CNN and Transformers were used.

   \item After pertaining, an end-to-end finetuning of the pre-trained model was done using x\% labeled data. Where x was either 1 or 10, or 100. When fine-tuned with x\% labeled data, the pre-trained model was then fine-tuned only on x\% data points with corresponding labels. 
\end{itemize}

\subsubsection{Augmentations}

\begin{itemize}

\item Resizing: Resize input images to 384×384 with bilinear interpolation.
\item Color jittering: change the brightness, contrast, saturation, and hue of an image or apply random perspective with a given probability. We set the degree of distortion to 0.2 (between 0 and 1) and use bilinear interpolation with an application probability of 0.3.
\item Color jittering or applying the random affine transformation of the image, keeping center invariant with degree 10, with an application probability of 0.3.
\item Horizontal and Vertical flip. Each with an application probability of 0.3.
\item Channel normalization with a mean (0.485, 0.456, 0.406) and standard deviation (0.229, 0.224, 0.225).

\end{itemize}

\subsubsection{Hyper-parameters}

\begin{itemize}
    \item Optimization: We use stochastic weighted averaging over Adam optimizer with learning rate (LR) set to 1e-3 for both CNN and vision transformer (ViT). This is a shift from SGD, which is usually used for CNNs.
    
    \item Learning Rate: Cosine annealing learning rate is used with 16 iterations and a minimum learning rate of 1e-6. Unless mentioned otherwise, this setup was trained over 100 epochs. These were then used as initialization for the downstream supervised learning. The standard batch size is 16.
    
\end{itemize}

\subsection{Supervised training}
\label{sup-train}
\subsubsection{Augmentations}
We use the same set of augmentations used in self-supervised pretraining.
\subsubsection{Hyper-parameters}
\begin{itemize}
    \item We use Adam optimizer with lr set to 3e-4 and a cosine annealing learning schedule.
    \item Since all medical datasets have class imbalance, we address it by using focal loss \cite{Lin2017FocalLF} as our choice of the loss function with the alpha value set to 1 and the gamma value to 2. In our case, it uses minimum-maximum normalized class distribution as class weights for focal loss.
    \item We train for 50 epochs. We also use a five epoch patience on validation loss to check for early stopping. This downstream supervised learning setup is kept the same for CNN and Transformers.
\end{itemize}

We repeat all the experiments with different seed values five times and then present the average results in all the tables.

\section{Miscellaneous}

\subsection{Description of Metrics}
After performing downstream fine-tuning on the four datasets under consideration, we analyze the CASS, DINO, and Supervised approaches on specific metrics for each dataset. The choice of this metric is either from previous work or as defined by the dataset provider. For the Autoimmune dataset, Dermofit, and Brain MRI classification datasets based in prior works, we use the F1 score as our metric for comparing performance, which is defined as $F1 = \frac{2*Precision*Recall}{Precision+Recall} = \frac{2*TP}{2*TP+FP+FN}$

For the ISIC-2019 dataset, as mentioned by the competition organizers, we used the recall score as our comparison metric, which is defined as $Recall = \frac{TP}{TP+FN}$

For the above two equations, TP: True Positive, TN: True Negative, FP: False Positive, and FN: False Negative.

\subsection{Limitations}

Although CASS' performance for larger and non-biological data can be hypothesized based on inferences, a complete study on large-sized natural datasets hasn't been conducted. In this study, we focused extensively on studying the effects and performance of our proposed method for small dataset sizes and in the context of limited computational resources. Furthermore, all the datasets used in our experimentation are restricted to academic and research use only. Although CASS performs better than existing self-supervised and supervised techniques, it is impossible to determine at inference time (without ground-truth labels) whether to pick the CNN or the Transformers arm of CASS.

\subsection{Potential negative societal impact}

The autoimmune dataset is limited to a geographic institution. Hence the study is specific to a disease variant. Inferences drawn may or may not hold for other variants.
Also, the results produced are dependent on a set of markers. Medical practitioners often require multiple tests before finalizing a diagnosis; medical history and existing health conditions also play an essential role. We haven't incorporated the meta-data above in CASS. Finally, application on a broader scale - real-life scenarios should only be trusted after clearance from the concerned health and safety governing bodies.
\end{document}